\documentclass{article}
\usepackage{spconf,amsmath,graphicx}
\usepackage{amsmath}
\usepackage{setspace}
\usepackage{multirow}
\usepackage{amsxtra, todonotes, soul}
\usepackage{cite}
\usepackage{threeparttable}
\usepackage[font=footnotesize]{subcaption}
\usepackage[font=footnotesize]{caption}
\usepackage{balance}

\title{HDR Image Watermarking using \\Saliency detection and quantization index modulation}

\name{Ahmed Khan$^1$, Minoru Kuribayashi$^2$, KokSheik Wong$^1$, Vishnu Monn Baskaran$^1$\vspace{-3mm}
\address{$^1$School of Information Technology, Monash University Malaysia, Malaysia.\\
	$^2$Graduate School of Natural Science and Technology, Okayama University, Japan. \vspace{-2mm}
}
}

\begin{document}

\maketitle

\begin{abstract}
	High-dynamic range (HDR) images are circulated rapidly over the internet with risks of being exploited for unauthorized usage. 
	To protect these images, some HDR image-based watermarking (HDR-IW) methods were put forward.
	However, they inherited the same problem faced by conventional IW methods for standard dynamic range (SDR) images, where only trade-offs among conflicting requirements are managed instead of simultaneous improvement. 
	In this paper, a novel saliency (eye-catching object) detection based trade-off independent HDR-IW is proposed, to simultaneously improve robustness, imperceptibility and payload.
	First, the host image goes through our proposed salient object detection model to produce a saliency map, which is, in turn, exploited to segment the foreground and background of the host image. 
	Next, the binary watermark is partitioned into the foregrounds and backgrounds using the same mask and scrambled using a random permutation algorithm. 
	Finally, the watermark segments are embedded into selected bit-plane of the corresponding host segments using quantized indexed modulation.
	Experimental results suggest that the proposed work outperforms  state-of-the-art methods in terms of improving the conflicting requirements.
\end{abstract}

\begin{keywords}
	Watermarking, Saliency detection, HDR images, High capacity, Robustness
\end{keywords}

\section{Introduction}
High dynamic range (HDR) imaging 
allows for better image quality and visual perception 
because it can represent a wider range of illumination in comparison to standard dynamic range (SDR) imaging~\cite{xue2011bilateral}. 
Consequently, HDRI is deployed for various applications, including computer graphics, computer vision, broadcasting, surveillance, to name a few~\cite{xue2011watermarking}.
Often these images are of commercial, political, educational, or sentimental value.
Hence they must be secured from unauthorized usage or illegal copying during transmission or in storage. 
Image watermarking (IW) is an art and science that injects copyright data into an image (host) to guard the host image.
The inserted data can be retrieved later from the potentially modified image for copyright claim, authentication, tampering detection, etc.~\cite{solachidis2013hdr1,guerrini2011high}.

\begin{figure}[t!]
	\centering
	\includegraphics[width=1\linewidth]{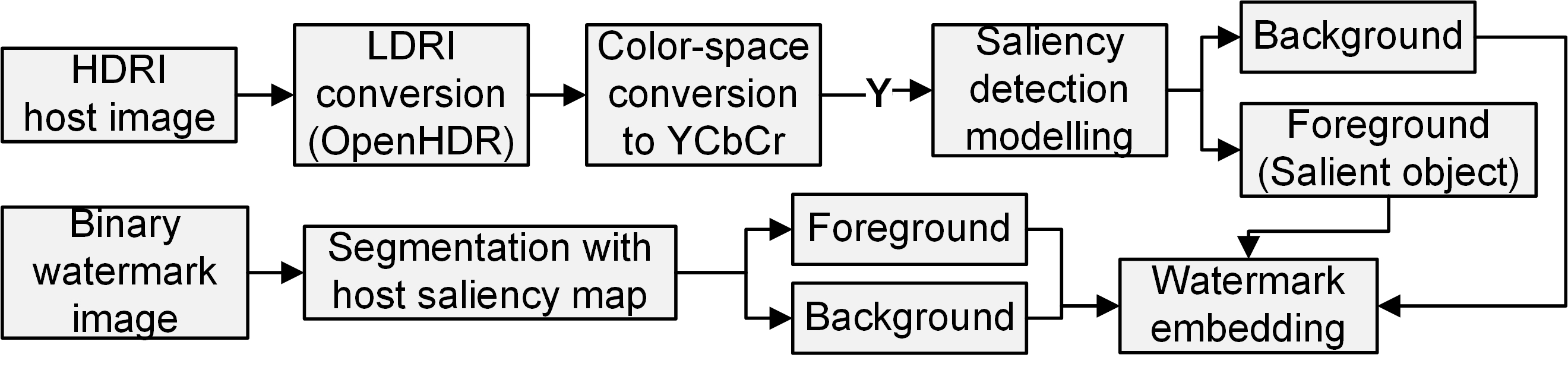}\vspace{-1mm}
	\caption{Proposed framework of HDR-IW method based on salient object detection. Detailed flow-diagrams of the proposed saliency detection model and watermark embedding procedure are shown in Fig.~\ref{Fig::model2} and \ref{Fig::modelIW}, respectively. ~\label{Fig::model} \vspace{-3mm}} 
\end{figure}

In the case of HDR-IW, various 
state-of-the-art (SOTA) IW methods are proposed, but often only a trade-off among capacity, quality, and robustness is achieved. Huang et al.~\cite{huang2021novel} designed a zero-bit IW method to protect the HDR image using dual-tree complex wavelet transform (DCWT) and Quaternion Fast Fourier transform (QFFT). 
First, DCWT and QFFT are sequentially applied to each color channel. 
Next, singular value decomposition (SVD) is applied to each LL subband, and finally, the watermark is embedded into the diagonal of the S matrices.
Perez et al.~\cite{perez2020watermarking} presented an HDR-IW method using Luma Variation Tolerance.
Specifically, the scrambled watermark is embedded into selected HDRI regions with less luma variation, to maintain the image quality.
Maiorana et al.~\cite{maiorana2016high, maiorana2013robust} and Shiju et al.~\cite{shiju2012performance} designed an HDR-IW system primarily based on Guerrini et al.'s method~\cite{guerrini2011high}. 
However, the watermarked image quality is better due to the utilization of the radon-DCT. 
Autrusseau et al.~\cite{autrusseau2013non} proposed an HDR-IW to improve the robustness feature. 
First, the RGB channels are extracted from the HDRI using the normalization process. 
Subsequently, DWT is applied to each color channel to obtain the LH sub-band for the watermark embedding using $\alpha$-blending.

In analyzing the aforementioned methods, it is evident that capacity is sacrificed in an attempt to improve the other two aspects of quality and robustness~\cite{khan2022-MTAP,maiorana2016high,maiorana2013robust,shiju2012performance,maiorana2016multi,solachidis2013hdr2,bai2018novel,autrusseau2013non,bakhsh2018robust,luo2019robust,huang2021novel,du2022robust,anbarjafari2018imperceptible}. To simultaneously achieve high embedding capacity, image quality, and robustness, one approach is to leverage on different saliency regions of an image for a watermarking process~\cite{khan2022-MTAP,khan2022-APSIPA}.
This can be achieved by using salient object detection (SoD) to extract the salient regions of an image to embed the watermark. With this approach, the critical regions of an image are used to embed the watermark without compromising the apparent visual quality. While subjectively reducing any observed visual distortions by the human eye, the IW can be made more resilient against different attacks. Motivated by this approach, we propose a novel framework for a reliable and invisible HDR-IW using SoD (see Fig~\ref{Fig::model} and Fig~\ref{Fig::model2}). The proposed method ensures that the quality of the watermarked image remains intact while protecting the watermark (i.e., by providing an extra layer of security).



\section{Methodology}
First, the host HDR image is tone-mapped 
to an SDR image $H$. 
From $H$, the foreground $H_f$ and background $H_b$ are determined using our proposed SoD-Net method (see Fig~\ref{Fig::model2}).
Based on the features of $H$, the watermark $W$ is then divided into two partitions $W_f$ and $W_b$ such that $W = W_f \bigcup W_b$, and both $W_f$ and $W_b$ are shuffled before being embedded into $H$ (see Fig.~\ref{Fig::model}). 
Details are provided in the following subsections.

\subsection{Res2Net based Salient Object Detection}
\begin{figure}[t!]
	\centering
	\includegraphics[width=1\linewidth]{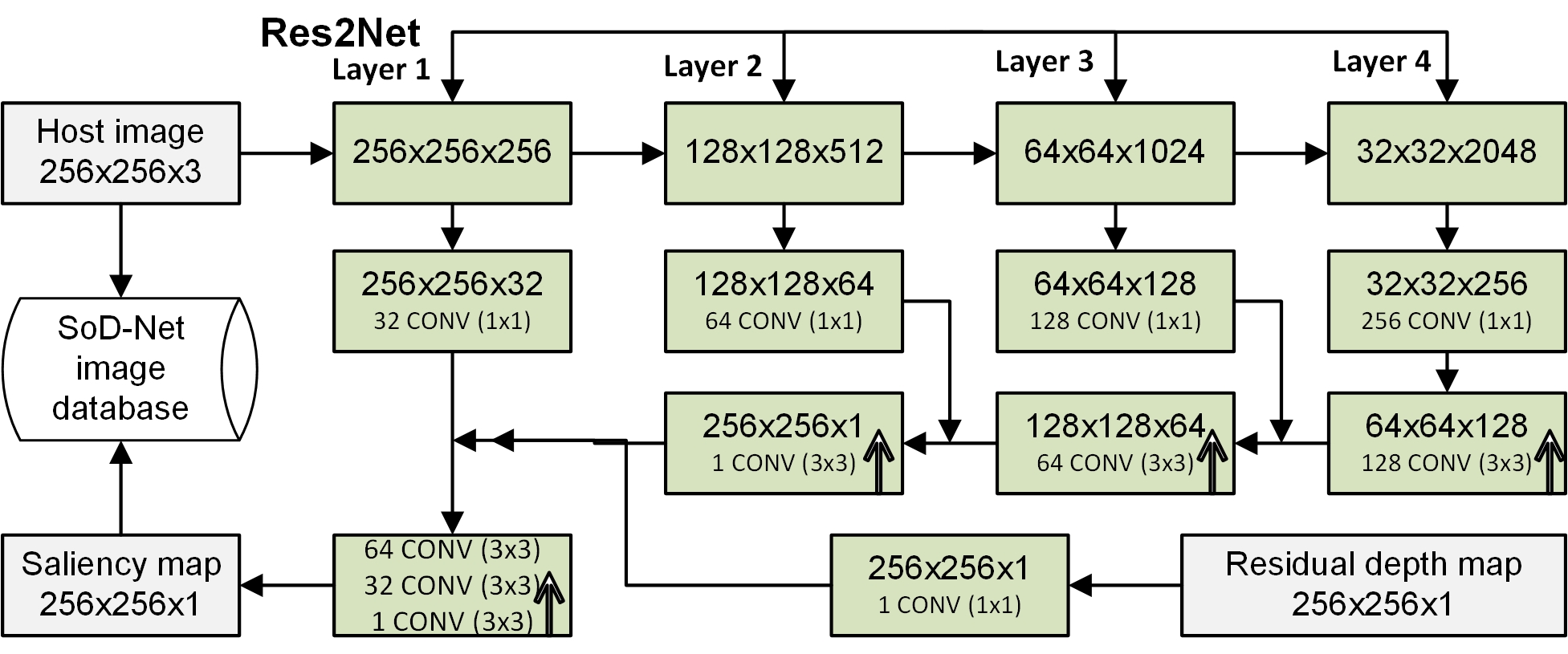}
	\caption{Proposed framework of SoD-Net  \label{Fig::model2} \vspace{-3mm}} 
\end{figure}
SoD-Net aims to detect salient objects in $H$ (see Fig.~\ref{Fig::model2}). 
First, HDR to SDR conversion is performed using normalized tone-mapping operation $\tau = \lceil \log_2[\max(c)+128]\rceil$
to get $H'_{c} = \lfloor \frac{c\cdot 256}{2^{(\tau-128)}}\rfloor$ and $H''_{c} = \lfloor \frac{H'_{c}+0.5+2^{(\tau-128)}}{256} \rfloor$,
where $c \in \{R,G,B\}$.
Here, $\{H''_{c},H'_{c}\}$ are the corresponding color channels $c$ in SDR and HDR. 
Next, the color and edge features are extracted from $H'$ using Res2Net~ \cite{wang2022res2fusion}. 
In SoD-Net, to verify that the output of Res2Net is an image, we eliminate all the completely linked layers.  
We remove the initial max pooling layer and, we set CONVs stride to 1 (instead of 2) to avoid excessive down-sampling ($\downarrow$).
As a result, sparse feature loss is avoided, while retaining the saliency details. 

Res2Net offers four outputs through the $\downarrow$ phases, namely, low-level features extracted at Layer 1, middle-level features at Layers 2 and 3, and sparse features at Layer 4, which we receive at $\downarrow$ phase.
One crucial use of $1\times1$ CONVs is as a dimension reduction module to eliminate computational constraints that would restrict the size of SoD-Net. 
It expands our networks' depth and width without suffering performance degradations. 
The number of channels is then reduced to 1/8 of the original number using four $1\times1$ CONVs.
The downsizing is performed because the original data is large in size, hence consuming more time for training and testing. 
Similarly, to create the depth saliency map, we erase the first max pooling layers, set the CONVs stride to 1, and utilize $3\times3$ CONVs. 
The depth error in non-salient objects causes the disparity map's nearest objects to have either the lowest or highest intensities.
These objects have the potential to deceive neural networks (NN) when NN models seek for salient objects in depth maps.

\begin{figure*}[t!]
	\begin{center}
		\begin{minipage}[t]{.16\linewidth}
			\centering
			\includegraphics[width=1\textwidth, height=2.8cm]{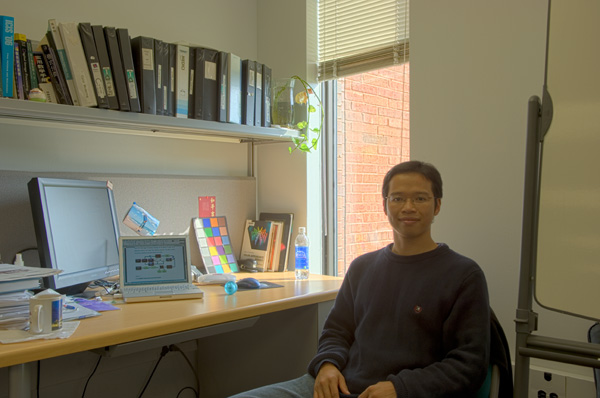}
			\subcaption{Host Image, $H$}
		\end{minipage}
		\begin{minipage}[t]{.16\linewidth}
			\centering
			\includegraphics[width=1\textwidth, height=2.8cm]{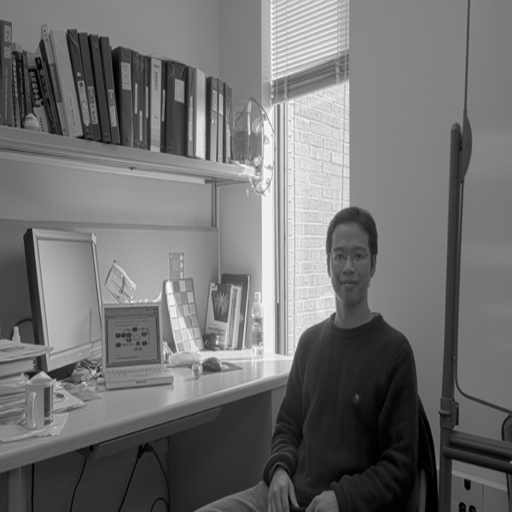}
			\subcaption{Luma, $Y$}
		\end{minipage}
		\begin{minipage}[t]{.16\linewidth}
			\centering
			\includegraphics[width=1\textwidth, height=2.8cm]{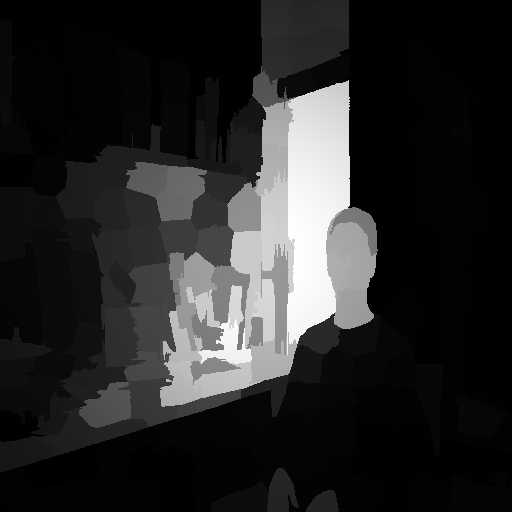}
			\subcaption{Saliency Map, $S$}
		\end{minipage}
		\begin{minipage}[t]{.16\linewidth}
			\centering
			\includegraphics[width=1\textwidth, height=2.8cm]{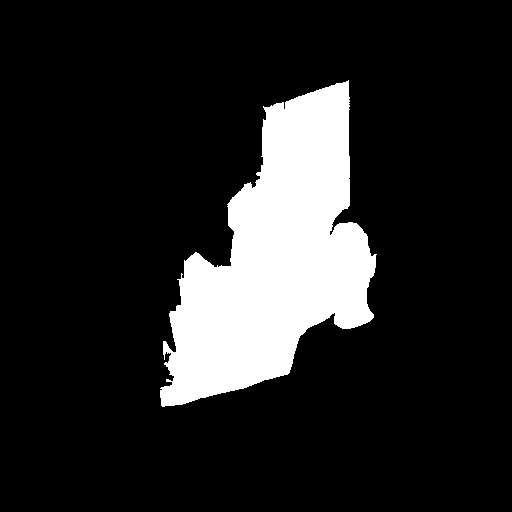}
			\subcaption{Binary Mask, $S^{B}$}
		\end{minipage}
		\begin{minipage}[t]{.16\linewidth}
			\centering
			\includegraphics[width=1\textwidth, height=2.8cm]{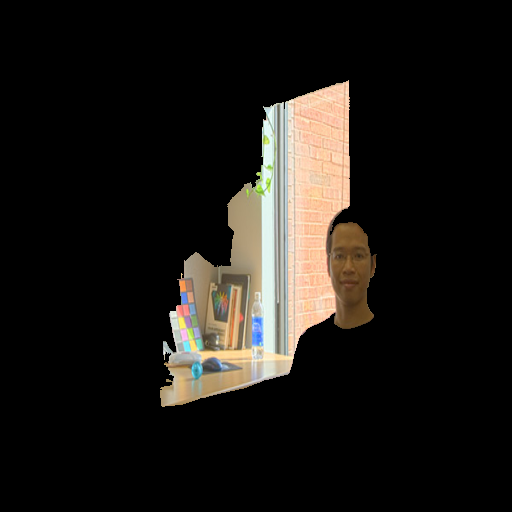}
			\subcaption{Salient Object, $H_f$}
		\end{minipage}
		\begin{minipage}[t]{.16\linewidth}
			\centering
			\includegraphics[width=1\textwidth, height=2.8cm]{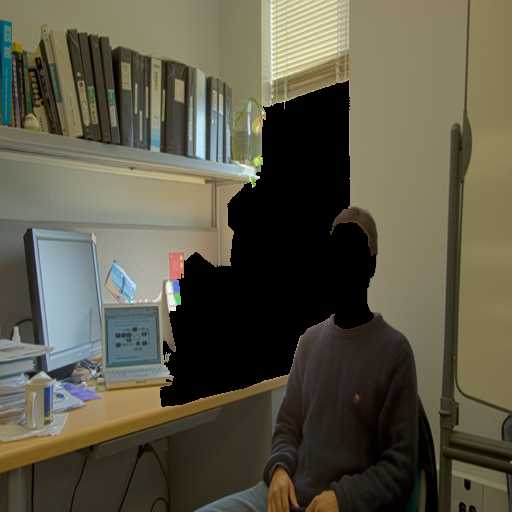}
			\subcaption{Background, $H_b$}
		\end{minipage}
		\begin{minipage}[t]{.16\linewidth}
			\centering
			\includegraphics[width=1\textwidth, height=2.8cm]{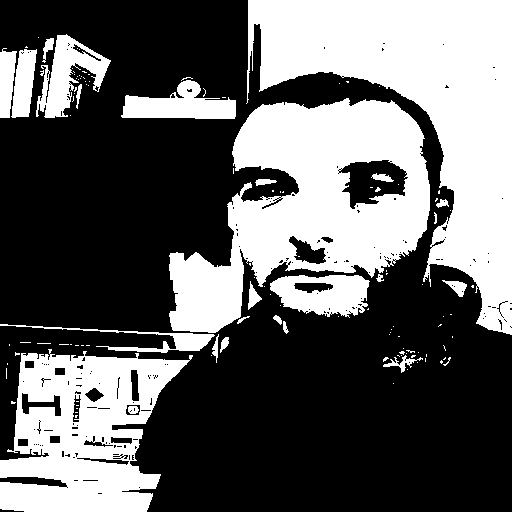} 
			\subcaption{Watermark, $W$}
		\end{minipage}
		\begin{minipage}[t]{.16\linewidth}
			\centering
			\includegraphics[width=1\textwidth, height=2.8cm]{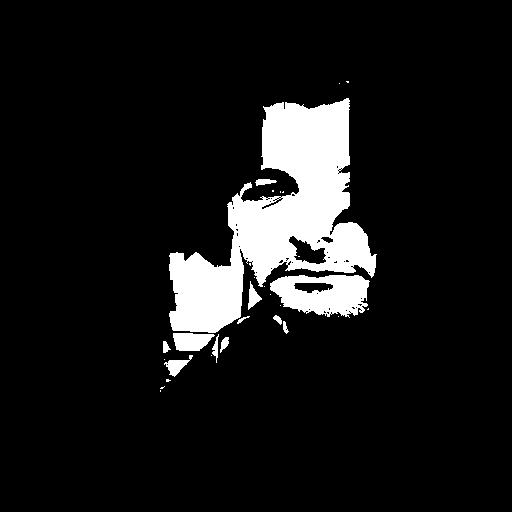}
			\subcaption{$W_f=W\wedge S^{B}$}
		\end{minipage}
		\begin{minipage}[t]{.16\linewidth}
			\centering
			\includegraphics[width=1\textwidth, height=2.8cm]{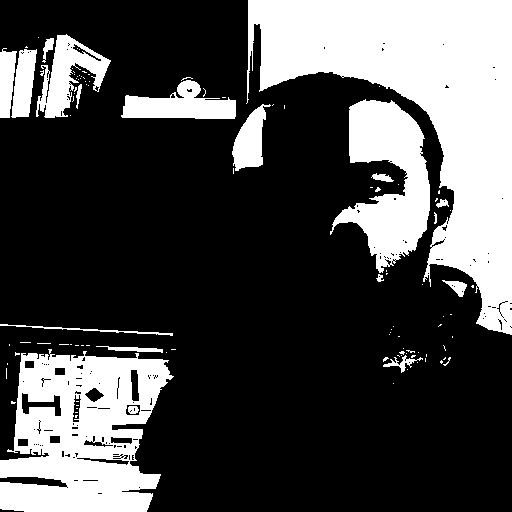}
			\subcaption{$W_b=W\wedge (1-S^{B})$}
		\end{minipage}
		\begin{minipage}[t]{.16\linewidth}
			\centering
			\includegraphics[width=1\textwidth, height=2.8cm]{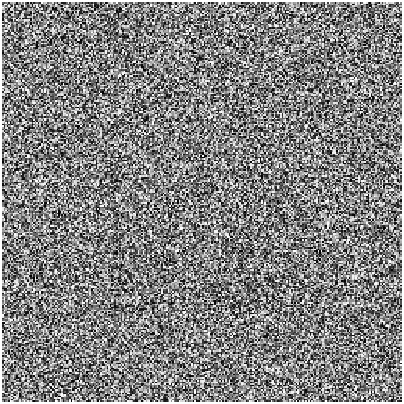}
			\subcaption{Encrypted, $W^e$}
		\end{minipage}
		\begin{minipage}[t]{.16\linewidth}
			\centering
			\includegraphics[width=1\textwidth, height=2.8cm]{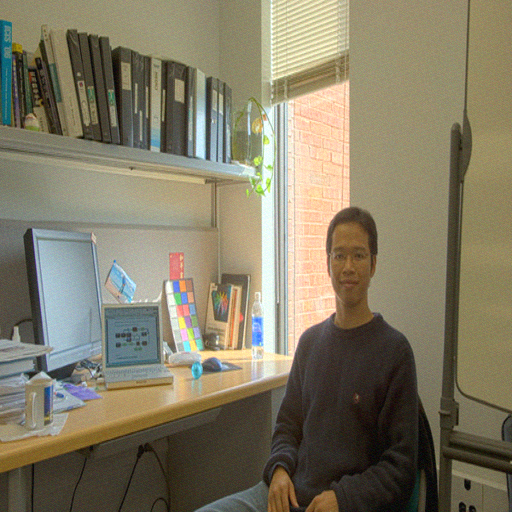}
			\subcaption{Watermarked, $H^W$}
		\end{minipage}
		\begin{minipage}[t]{.16\linewidth}
			\centering
			\includegraphics[width=1\textwidth, height=2.8cm]{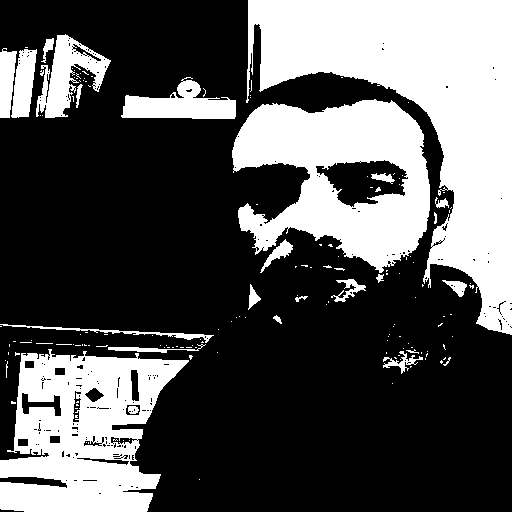}
			\subcaption{Extracted, $W'$}
		\end{minipage}\vspace{-1mm}
		\caption{Intermediate images produced by proposed image watermarking method }\label{Fig::example}
	\end{center} \vspace{-5mm}
\end{figure*}

As a result, a single-modal saliency map that simply includes depth information is flawed.
By building a hybrid saliency map using the additive RGB and depth information, we can get rid of depth inaccuracy.
Specifically, the weighted saliency map created by combining the RGB and depth modes contains the likelihood information of each pixel 
to be part of a certain object. 
We design a decoder network that consists of $3\times3$ CONVs and up-sampling ($\uparrow$) to produce the cumulative saliency map.
We concatenate the lower-level and current features following each $\uparrow$. 
The decoder block $D^{n}$ is as follows \vspace{-2mm}
\begin{equation}
	S = \bigcup D^{n} = \sum_{i=1}^{k} \uparrow (CONV(D_{i}^{n-1}\bigoplus d_{i}^{n}.\Theta)), 
\end{equation}
where $i$ refers to the $i^{th}$ feature channel, $CONV$ is the convolution with value $\Theta$, and $\bigoplus$ is the \emph{concatenation} operator. 
The binary mask $S^B$ is produced by thresholding $S$~\cite{khan2022-MTAP}. 
The loss function is $L(S,g) = \sum \sum (S \log g) + (1-S) \log(1-g)$, where $g$ is the ground truth.
Finally, $H_{b}$ is obtained by $H_{b} = H' \land (1-S^B)$. 
Likewise, $H_{f}$ is obtained by $H_{f} = H' \land S$. 
Subsequently, $W_f$ and $W_b$ are calculated (see Fig.~\ref{Fig::example}).
\begin{figure}[t!]
	\centering
	\includegraphics[width=1\linewidth]{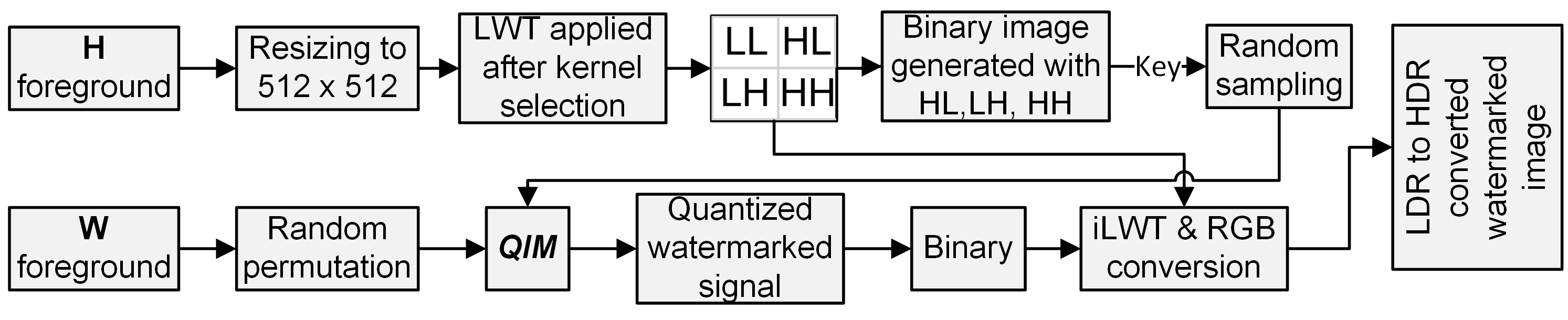}\vspace{-1mm}
	\caption{Proposed framework of our HDR-IW design  \label{Fig::modelIW} \vspace{-2mm}} 
\end{figure}

\subsection{QIM based Watermark Embedding}

	We embed the divided watermark $\{W_f, W_b\}$ correspondingly into $\{H_f, H_b\}$ after shuffling them according to a secret key (see Fig.~\ref{Fig::modelIW}). 
	Details are provided in the following steps.
	
	\textbf{Step 1:} First, $\{H_f, H_b\}$ are resized (cubic interpolation) to $512 \times 512$, followed by the $YC_bC_r$ conversion. 
	Next, lifting wavelet transform (LWT) using \textit{Daubechies cdf2.2-kernel} is applied to $H_f$ of $Y$ to produce the sub-bands $[LL,HL,LH,HH]_{\downarrow}$. Similarly, LWT using \textit{Haar-kernel} is applied to $H_b$ of $Y$. 
	Note that we apply the same method for both $\{H_f, H_b\}$ to embed $\{W_f, W_b\}^e$, which is of size $H/2$. 
	
	\textbf{Step 2:} Subsequently, the shuffled binary $\{W_f, W_b\}^e$ is generated by applying random permutation to $\{W_f, W_b\}$.
	Next, a binary image $B_{\kappa, (i,j)}^n \in \{0,1\}$ is constructed from the $n$-th bit plane of $\kappa\in \{HL,LH,HH\}$ sub-band. 
	In order to generate the sampled host signal $\varphi_{(i,j)}$, a random sampling operation is applied to $B_{\kappa, (i,j)}^n$ according to a secret key.
	
	\textbf{Step 3:} The quantization modulation operation is applied to embed $W_{(i,j)}^e$, which is the $(i,j)$-th element of $\{W_f, W_b\}^e$, in $\varphi_{(i,j)}$.
	The watermarked signal $\varphi^W_{(i,j)}$ is computed as: 
	\begin{equation} 
		\label{QIM}
		\begin{array}{ll}
			\varphi^W_{(i,j)} & = \textrm{QIM}(\varphi_{(i,j)}, W_{(i,j)}^e, \Delta)\\
			& = \left\{
			\begin{array}{ll}
				\Delta \cdot \tilde{\varphi}_{(i,j)} & \text{if} \hspace{1mm} W_{(i,j)}^e = \tilde{\varphi}_{(i,j)} \pmod 2;\\  
				\Delta \cdot \tilde{\varphi}_{(i,j)}+1 & \text{otherwise,}
			\end{array}\right.
		\end{array}
	\end{equation}
	where $\Delta$ is the quantization step and $\tilde{\varphi}_{(i,j)}=\lfloor \frac{\varphi_{(i,j)}}{\Delta} \rfloor$. Next, $\tau$ is embedded into $LL$ by using Eq.~(\ref{QIM}) for synchronized robust tone-mapping.
	
	\textbf{Step 4:} The binary form of $\varphi^W_{(i,j)}$ is produced in order to combine it with the binary $B_{\kappa, (i,j)}^n$ to perform the smooth inverse operation. Finally, inverse LWT 
	followed by \textit{$YC_bC_r$} to \textit{RGB} conversion are computed to achieve $H'^W$. 
	Finally, inverse tone-mapping is applied to $H'^W$ to produce the HDR watermarked $H^W$. 
	Intermediate results are shown in Fig.~\ref{Fig::example}. 
	
	It is noteworthy that $W$ can be extracted as $W'$ by reversing the embedding steps. 
	In addition, the watermark is extracted in the SDR domain, and hence robustness against tone-mapping is important and is of our interest.

	\section{Experiments and Results}
	\begin{figure}[t!]
		\begin{center}
			\begin{minipage}[t]{0.32\linewidth}
				\includegraphics[width=1\textwidth, height=2.8cm]{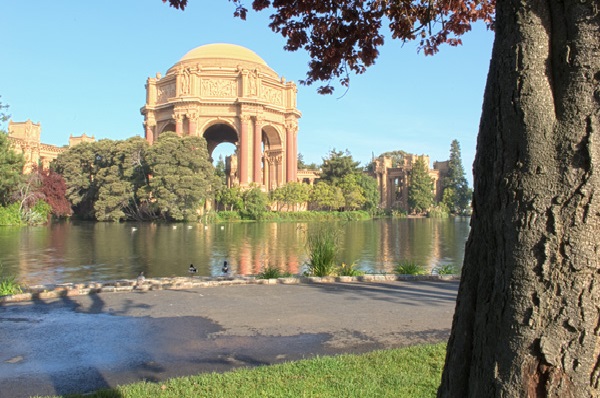}
				\subcaption{Exploratorium}
			\end{minipage}
			\begin{minipage}[t]{0.32\linewidth}
				\includegraphics[width=1\textwidth, height=2.8cm]{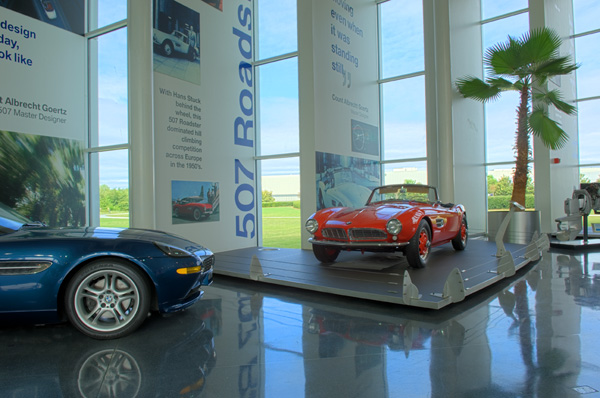}
				\subcaption{BMW 507}
			\end{minipage}
			\begin{minipage}[t]{0.32\linewidth}
				\includegraphics[width=1\textwidth, height=2.8cm]{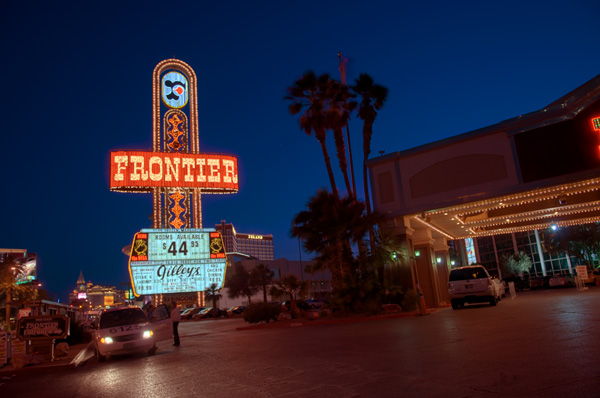}
				\subcaption{Frontier}
			\end{minipage}\vspace{-3mm}
			\caption{Additional host images for experiments\vspace{-5mm}} 
			\label{Fig::AdditionalHost} 
		\end{center}
	\end{figure}
	\begin{table}[t!]
		\setstretch{0.98}
		\setlength{\tabcolsep}{.41em}
		\centering
		\caption{Analyzing quality of $W$ and $H^W$ for our and SOTA methods\vspace{-1mm}\label{Table::PSNRSSIMNC}}
		\begin{footnotesize}
			\begin{tabular}{c|cccccc}
				Image   &   Proposed  &   \cite{gao2020reversible} &   \cite{bai2021reversible} &   \cite{xue2011bilateral} &
				\cite{du2022robust} &   
				\cite{huang2021novel} \\ \hline   
				& \multicolumn{4}{c}{Man - Fig.~\ref{Fig::example}(a)} \\ \hline
				PSNR (dB)   &   \textbf{59.85}   &   49.60 & 48.97&   50.43 & 46.22 & 45.04\\
				SSIM        &   \textbf{0.9999} &   - &   0.9874 & - & -&0.9867\\
				NC for $W'$   &   0.9999  &  0.9865 & \textbf{1} & \textbf{1} & 0.9820&0.9761\\ \hline 
				& \multicolumn{4}{c}{Exploratorium - Fig.\ref{Fig::AdditionalHost}(a)} \\ \hline
				PSNR (dB)   & \textbf{60.22} & 48.33 &   47.30 & 49.46 & 45.20& 45.37\\
				SSIM        & \textbf{0.9999} &  - &   0.9869 & - & -&0.9866\\
				NC for $W'$   &  0.9999 &  0.9835  &   \textbf{1} & 0.9989 & 0.9630&0.9761\\ \hline
				& \multicolumn{4}{c}{BMW 507 - Fig.\ref{Fig::AdditionalHost}(b)} \\ \hline
				PSNR (dB)   & \textbf{59.92} & 48.78 &   48.11 & 48.99 & 45.32& 45.11\\
				SSIM  & \textbf{0.9999} & - &   0.9927 & - & -& 0.9892\\
				NC for $W'$   & 0.9999  &  0.9769 &   0.9908 & \textbf{0.9999} &0.9750&0.9761\\ \hline
				& \multicolumn{4}{c}{Frontier - Fig.\ref{Fig::AdditionalHost}(c)} \\ \hline
				PSNR (dB)   & \textbf{59.81} &  49.78 &   46.30 & 49.08 & 46.05& 45.19 \\
				SSIM  & \textbf{0.9999}  & - &  0.9811 & -& - &0.9890\\
				NC for $W'$   & 0.9999 &  0.9862 &  0.9992 & \textbf{0.9999} &0.9790&0.9761 \\ \hline
			\end{tabular}\vspace{-3mm}
		\end{footnotesize}
	\end{table}
	\begin{figure*}
		\begin{minipage}{.24\textwidth}
			\includegraphics[width=1\textwidth, height=1in]{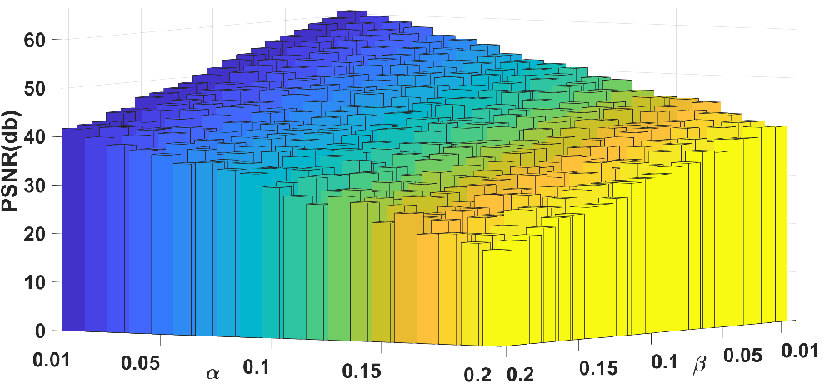}
			\subcaption{PSNR for $H^W$
			}
		\end{minipage} 
		\begin{minipage}{.24\textwidth}
			\includegraphics[width=1\textwidth, height=1in]{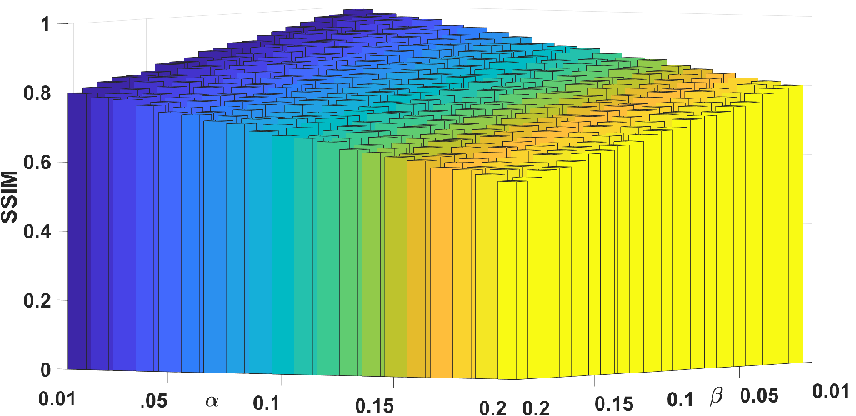}
			\subcaption{SSIM for, $H^W$ }
		\end{minipage}
		\begin{minipage}{.24\textwidth}
			\includegraphics[width=1\textwidth, height=1in]{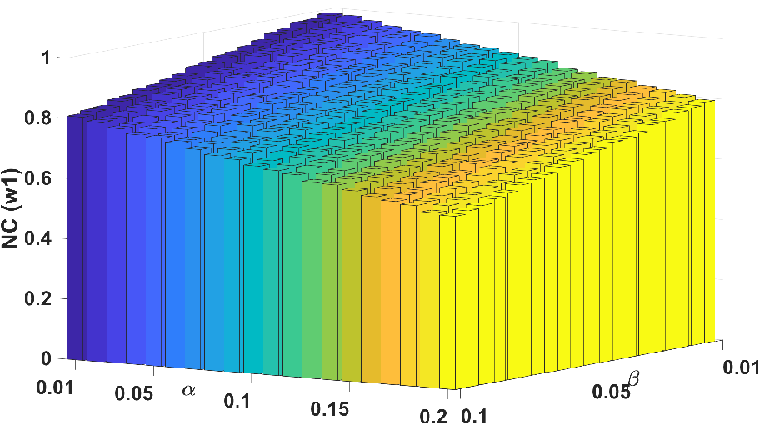}
			\subcaption{NC for $W_1$}
		\end{minipage} 
		\begin{minipage}{.24\textwidth}
			\includegraphics[width=1\textwidth, height=1in]{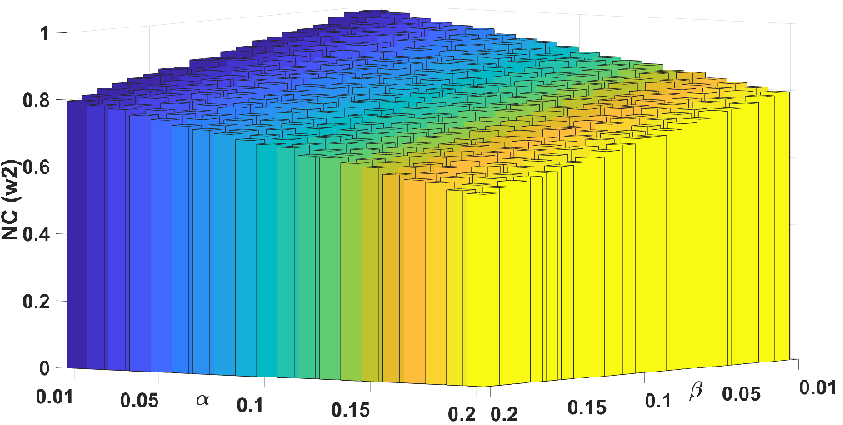}
			\subcaption{NC for $W_2$}
		\end{minipage}  \vspace{-1mm}
		\caption{The graphs of image quality and NC values for various combinations of $\alpha$ and $\beta$ values. The NC value shown is the average NC value for the seven attacks mentioned in Table~\ref{Table::ResultAttack}. 
			\label{graphsanalysis}}\vspace{-3mm}
	\end{figure*}
	Our SoD-Net HDR-IW method is implemented in Matlab 2020 (Windows 10) running on a PC with an AMD Ryzen 5 3500U, Radeon Vega Gfx, and 16GB-RAM.
	For experiment and comparison purposes, we used the RIT-MCSL~\cite{bai2021reversible} and MSRA datasets\footnote{For each SDR image, we produce 6 images each of different exposure, which are then combined using Durand's inverse TMO method~\cite{shi2021novel}. }~\cite{singh-MTAP-2020}.
	Additionally, we set the background to be slightly hazy ($\alpha=0.04$) and salient object to be entirely invisible ($\beta=0.02$). 
	Visual results are shown in Fig.~\ref{Fig::example}. 
	
	The PSNR and SSIM of $H^W$ are evaluated \cite{anbarjafari2018imperceptible}.
	Results suggest that $H^W$ produced by our method is of high quality, i.e., $>59$dB in terms of PSNR, and it is consistently higher 
	in comparison to SOTA methods (see Table~\ref{Table::PSNRSSIMNC}).
	The SSIM values also indicate that our watermarked images are of high quality, which agrees with the PSNR value. 
	
	Next the normalized correlation (NC)~\cite{huang2021novel} and $NC \in [0,1]$, is considered to assess the quality of the extracted $W'$. 
	\begin{equation} 
		NC=\frac{\sum_{(i,j)}(W_{(i,j)}-\mu_{W})(W^\prime_{(i,j)}-\mu_{W^\prime})}{\sqrt{\sum_{(i,j)}(W_{(i,j)}-\mu_{W})^{2}}\sqrt{\sum_{(i,j)}(W^\prime_{(i,j)}-\mu_{W^\prime})^{2}}}
	\end{equation}
	where, $\mu_W$ is average value of the $W$. 
	As expected, for the case of not undergoing any attack, the NC value is high as reported in  Table~\ref{Table::PSNRSSIMNC} (see row NC for $W'$). 
	The results here are collected using Durand's TMO.  
	
	To evaluate robustness, $H^W$ is subjected to the mean and median filterings, affine shearing $[1,0,0;0.5,1,0;0,0,1]$, additive noise $(0.01)$, rotation of $45^o$, cropping with $[75:424, 68:424]$, and JPEG compression to assess the robustness of our HDR-IW method.
	Here, all attacks take place in the tone-mapped SDR images produced by using Durand's TMO. 
	The results are recorded in Table~\ref{Table::ResultAttack}. 
	Note that two cases are considered here, namely, using the proposed SoD-Net approach ($NC_{W_1}$) and embedding equally throughout the host image ($NC_{W_2}$). 
	\begin{table}[t!]
		\setlength{\tabcolsep}{.24em}
		\centering
		\caption{Results NC after applying watermark attack\label{Table::ResultAttack}} 
		\begin{footnotesize}
			\begin{tabular}{l|c|c|c|c|c|c|c} 
				Attack & $NC_{W_1}$ & $NC_{W_2}$ & 
				\cite{gao2020reversible} &   
				\cite{bai2021reversible} &   
				\cite{shi2021novel} &
				\cite{du2022robust} &   
				\cite{huang2021novel} \\[0.5mm] \hline
				$^1$Mean-filter & 0.9861 & 0.9699 & \textbf{0.9872} & 0.9718 & 0.9748&- & 0.9710 \\
				$^2$Median-filter & 0.9848 & 0.9683 & \textbf{0.9911} & - & - &0.9840&-\\
				$^3$Shearing & \textbf{0.9864} & 0.9675 & - & - & 0.9513 &0.9658&- \\
				$^4$Noise & \textbf{0.9920} & 0.9792 & 0.8931 & 0.9680 & 0.9905 &0.9850&0.9178\\
				$^5$Rotation & \textbf{0.9791} & 0.9518 & 0.9090 & 0.9055 & 0.6335 &0.9498&0.8799\\
				$^6$Cropping & \textbf{0.9893} & 0.9681 & - & 0.9552 & 0.9602 &-&0.9685 \\
				$^7$JPEG-Comp & 0.9672 & 0.9446 & - & \textbf{0.9688} & 0.9559 &-&0.9220 \\\hline
			\end{tabular}
		\end{footnotesize} \vspace{-3mm}
	\end{table}
	First, results show that $NC_{W_1} > NC_{W_2}$ for all attacks.
	This outcome suggest that embedding $W$ into salient regions of the host image leads to better robustness against attacks. 
	It is because $\{W'_f, W'_b\}$ are embedded into different regions (details) of $H$ by using different embedding methods. 
	Simply said, $W$ is spread in different regions, “\textit{differently}”. 
	Besides robustness, SoD based embedding also ensures the imperceptibility feature. 
	Hence, this leads to concurrent improvement in watermarking conflicting requirement.
	On the other hand, when embedding $W$ evenly throughout the image (without SoD), $H^W$ may undergo the attacks evenly and hence attaining low robustness. 
	
	When compared to the SOTA methods, the proposed SoD-Net HDR-IW method achieves the high robustness for four out of seven attacks.
	It should be noted that, although our method does not achieve the highest NC value for the Mean-filter, Median-filter, and JPEG-Comp attacks, our NC values are only marginally below the best performing methods (i.e.,\cite{gao2020reversible,bai2021reversible}, respectively).
	Our HDR-IW  performs better against noise attack. 
	This result may have resulted from $W$ being shuffled before embedding into $H$, which spreads $W$, as opposed to the typical SOTA approaches that do not encrypt the $W$ prior to embedding. 
	In addition, 
	for the same host image $H$, the SOTA methods can embed less than 1 bit per pixel (bpp) (e.g., \cite{gao2020reversible,du2022robust} can only embed $128 \times 128$ and $32\times32$ bits, respectively), while our proposed method can diffuse 1 bit of $W$ into 1 pixel of $H$, i.e., 1 bpp.
	Interestingly, despite embedding significantly more data, our method produces $H^W$ of the highest quality when compared to the SOTA methods.
	
	Last but not least, we also analyzed the quality of the watermarked HDR image $H^W$ and the extracted watermark $W'$ with different $(\alpha, \beta)$ values as shown in Fig.~\ref{graphsanalysis}.
	Results suggest that increment in $\alpha$ and $\beta$ produces lower quality $H^W$, which eventually generates low quality $W'$, and vice versa.
	The observation is made for robustness. 
	However, for security purposes, one should avoid using the same $(\alpha,\beta)$ combination all the time. 
	
	All in all, we conclude that our HDR-IW method outperforms SOTA HDR-IW methods in terms of imperceptibility, robustness, and payload capability, hence achieving simultaneous improvement, viz., trade-off independence.
	
	\section{Conclusion}
	In this study, a novel trade-off independent HDR iamge watermarking based on saliency detection is
	proposed, which simultaneously achieves higher robustness, image quality and payload. 
	First, the HDR image is tone-mapped to an SDR image, which is then segmented into the background and foreground regions based on a mask produced by the proposed neural-network based SoD method called SoD-Net. 
	Prior to embedding, bit-plane of the watermark are partitioned into foregrounds and backgrounds using the same mask derived from the host and scrambled using the pseudo-random permutation algorithm. 
	Next, the watermark segments are embedded into the chosen bit-planes of the host image segments using the quantized indexed modulation. 
	Results suggest that our method is resilient to the non-malicious attacks (TMOs) and outperforms SOTA IW methods in terms of image quality, robustness, and payload. 
	
	As future work, we analyze the effect of different wavelets and decomposition levels on proposed HDR-IW method. 
	We also investigate the performance of the proposed HDR-IW method using different tone mapping operators.

	\balance
	\footnotesize \bibliographystyle{IEEEbib}
	\bibliography{refs}

\begin{thebibliography}{10}

\bibitem{xue2011bilateral}
Xinwei Xue, Masahiro Okuda, and Satoshi Goto,
\newblock ``Bilateral filtering based watermarking for high dynamic range
  image,''
\newblock in {\em 2011 International Symposium on Intelligent Signal Processing
  and Communications Systems (ISPACS)}. IEEE, 2011, pp. 1--5.

\bibitem{xue2011watermarking}
Xinwei Xue, Takao Jinno, Xin Jin, Masahiro Okuda, and Satoshi Goto,
\newblock ``Watermarking for hdr image robust to tone mapping,''
\newblock {\em IEICE Transactions on Fundamentals of Electronics,
  Communications and Computer Sciences}, vol. 94, no. 11, pp. 2334--2341, 2011.

\bibitem{solachidis2013hdr1}
Vassilios Solachidis, Emanuele Maiorana, Patrizio Campisi, and Francesco
  Banterle,
\newblock ``Hdr image watermarking based on bracketing decomposition,''
\newblock in {\em 2013 18th International Conference on Digital Signal
  Processing (DSP)}. IEEE, 2013, pp. 1--6.

\bibitem{guerrini2011high}
Fabrizio Guerrini, Masahiro Okuda, Nicola Adami, and Riccardo Leonardi,
\newblock ``High dynamic range image watermarking robust against tone-mapping
  operators,''
\newblock {\em IEEE Transactions on Information Forensics and Security}, vol.
  6, no. 2, pp. 283--295, 2011.

\bibitem{khan2022-MTAP}
Ahmed Khan and KokSheik Wong,
\newblock ``High payload watermarking based on enhanced image saliency
  detection,''
\newblock {\em Multimedia Tools and Applications}, pp. 1--19, 2022.

\bibitem{huang2021novel}
Jiangtao Huang, Shanshan Shi, Zhouyan He, and Ting Luo,
\newblock ``A novel zero watermarking based on dt-cwt and quaternion for hdr
  image,''
\newblock {\em Electronics}, vol. 10, no. 19, pp. 2385, 2021.

\bibitem{perez2020watermarking}
Karina~Ruby Perez-Daniel, Francisco Garcia-Ugalde, and Victor Sanchez,
\newblock ``Watermarking of hdr images in the spatial domain with
  hvs-imperceptibility,''
\newblock {\em IEEE Access}, vol. 8, pp. 156801--156817, 2020.

\bibitem{maiorana2016high}
Emanuele Maiorana and Patrizio Campisi,
\newblock ``High-capacity watermarking of high dynamic range images,''
\newblock {\em EURASIP Journal on Image and Video Processing}, vol. 2016, no.
  1, pp. 1--15, 2016.

\bibitem{maiorana2013robust}
Emanuele Maiorana, Vasileios Solachidis, and Patrizio Campisi,
\newblock ``Robust multi-bit watermarking for hdr images in the radon-dct
  domain,''
\newblock in {\em 2013 8th International Symposium on Image and Signal
  Processing and Analysis (ISPA)}. IEEE, 2013, pp. 284--289.

\bibitem{shiju2012performance}
KP~Shiju and P~Tamil Selvi,
\newblock ``Performance analysis of high dynamic range image watermarking based
  on quantization index modulation,''
\newblock in {\em 2012 International Conference on Power, Signals, Controls and
  Computation}. IEEE, 2012, pp. 1--5.

\bibitem{autrusseau2013non}
Florent Autrusseau and Dalila Goudia,
\newblock ``Non linear hybrid watermarking for high dynamic range images,''
\newblock in {\em 2013 IEEE International Conference on Image Processing}.
  IEEE, 2013, pp. 4527--4531.

\bibitem{maiorana2016multi}
Emanuele Maiorana and Patrizio Campisi,
\newblock ``Multi-bit watermarking of high dynamic range images based on
  perceptual models,''
\newblock {\em Security and Communication Networks}, vol. 9, no. 8, pp.
  705--720, 2016.

\bibitem{solachidis2013hdr2}
Vassilis Solachidis, Emanuele Maiorana, and Patrizio Campisi,
\newblock ``Hdr image multi-bit watermarking using bilateral-filtering-based
  masking,''
\newblock in {\em Image Processing: Algorithms and Systems XI}. SPIE, 2013,
  vol. 8655, pp. 29--40.

\bibitem{bai2018novel}
Yongqiang Bai, Gangyi Jiang, Hao Jiang, Mei Yu, Fen Chen, and Zhongjie Zhu,
\newblock ``Novel robust high dynamic range image watermarking algorithm
  against tone mapping,''
\newblock {\em KSII Transactions on Internet and Information Systems (TIIS)},
  vol. 12, no. 9, pp. 4389--4411, 2018.

\bibitem{bakhsh2018robust}
F~Yazdan Bakhsh and Mohsen~Ebrahimi Moghaddam,
\newblock ``A robust hdr images watermarking method using artificial bee colony
  algorithm,''
\newblock {\em Journal of Information Security and Applications}, vol. 41, pp.
  12--27, 2018.

\bibitem{luo2019robust}
Ting Luo, Gangyi Jiang, Mei Yu, Haiyong Xu, and Wei Gao,
\newblock ``Robust high dynamic range color image watermarking method based on
  feature map extraction,''
\newblock {\em Signal Processing}, vol. 155, pp. 83--95, 2019.

\bibitem{du2022robust}
Meng Du, Ting Luo, Haiyong Xu, Yang Song, and Chunpeng Wang,
\newblock ``Robust hdr video watermarking method based on saliency extraction
  and t-svd,''
\newblock {\em The Visual Computer}, vol. 38, no. 11, pp. 3775--3789, 2022.

\bibitem{anbarjafari2018imperceptible}
Gholamreza Anbarjafari and Cagri Ozcinar,
\newblock ``Imperceptible non-blind watermarking and robustness against tone
  mapping operation attacks for high dynamic range images,''
\newblock {\em Multimedia Tools and Applications}, vol. 77, no. 18, pp.
  24521--24535, 2018.

\bibitem{khan2022-APSIPA}
Ahmed Khan, KokSheik Wong, and Vishnu~Monn Baskaran,
\newblock ``Image watermarking based on saliency detection and multiple
  transformations,''
\newblock in {\em 2022 Asia-Pacific Signal and Information Processing
  Association Annual Summit and Conference (APSIPA ASC)}. IEEE, 2022, pp.
  1514--1518.

\bibitem{wang2022res2fusion}
Zhishe Wang, Yuanyuan Wu, Junyao Wang, Jiawei Xu, and Wenyu Shao,
\newblock ``Res2fusion: Infrared and visible image fusion based on dense
  res2net and double nonlocal attention models,''
\newblock {\em IEEE Transactions on Instrumentation and Measurement}, vol. 71,
  pp. 1--12, 2022.

\bibitem{gao2020reversible}
Xinyi Gao, Zhibin Pan, Erdun Gao, and Guojun Fan,
\newblock ``Reversible data hiding for high dynamic range images using
  two-dimensional prediction-error histogram of the second time prediction,''
\newblock {\em Signal Processing}, vol. 173, pp. 107579, 2020.

\bibitem{bai2021reversible}
Yongqiang Bai, Gangyi Jiang, Zhongjie Zhu, Haiyong Xu, and Yang Song,
\newblock ``Reversible data hiding scheme for high dynamic range images based
  on multiple prediction error expansion,''
\newblock {\em Signal Processing: Image Communication}, vol. 91, pp. 116084,
  2021.

\bibitem{shi2021novel}
Shanshan Shi, Ting Luo, Jiangtao Huang, and Meng Du,
\newblock ``A novel hdr image zero-watermarking based on shift-invariant
  shearlet transform,''
\newblock {\em Security and Communication Networks}, vol. 2021, 2021.

\bibitem{singh-MTAP-2020}
Surya~Kant Singh and Rajeev Srivastava,
\newblock ``A robust salient object detection using edge enhanced global
  topographical saliency,''
\newblock {\em Multimedia Tools and Applications}, pp. 1--18, 2020.

\end{thebibliography}
	
\end{document}